\documentclass[letterpaper, 10 pt, conference]{ieeeconf}  

\IEEEoverridecommandlockouts                              
                                                          
\let\labelindent\relax

\usepackage{amsmath}
\usepackage{gensymb}
\usepackage{times}
\usepackage{epsfig}
\usepackage{subfigure}
\usepackage[font=footnotesize,justification=justified]{caption}
\usepackage{graphicx}
\usepackage{amsmath, amsfonts, amssymb, amsthm, bm}
\usepackage{algpseudocode}
\usepackage{algorithm}
\usepackage{multicol}
\usepackage{multirow}
\usepackage{graphbox}
\usepackage[bookmarks=true]{hyperref}
\usepackage{float}
\usepackage{cite}

\usepackage{enumitem}

\usepackage{todonotes}

\begin{document}

\title{HOPE-Net: Hand - Object Pose Estimation in RGB Images}
\title{Understanding Manipulation through Hand - Object Pose Estimation in RGB Images}
\title{Understanding Manipulation Actions from\\Hand-Object Pose Estimation in RGB Images}
\title{{\bf Object-in-Hand Pose Estimation in RGB Images}}
\title{\LARGE \bf Understanding Hand-Object Interactions Through 6D Pose Estimation from RGB Images}
\title{\LARGE \bf Towards Understanding Manipulation Actions through\\Object-in-Hand Pose Estimation from RGB Images}
\title{\LARGE \bf Object-in-Hand Shape and Pose Estimation from RGB Images}
\title{\LARGE \bf Learning to Estimate Pose and Shape\\of Hand-Held Objects from RGB Images}

\makeatletter
\let\@oldmaketitle\@maketitle
\renewcommand{\@maketitle}{\@oldmaketitle
  \includegraphics[width=\linewidth]{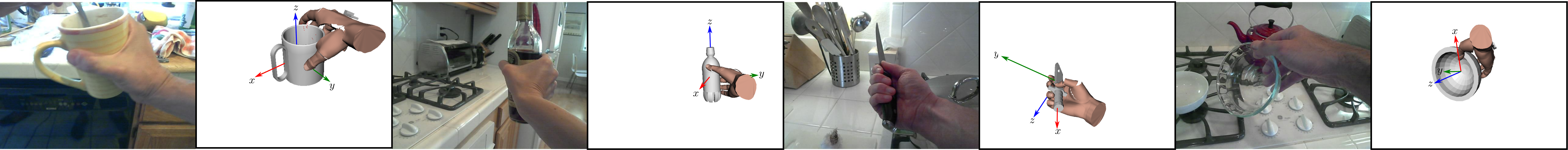}
  \captionof{figure}{Qualitative example results for estimating the 6D pose and the shape of hand-held objects from RGB images. We show four input-output pairs of our approach - one per object category considered in this paper: \emph{mug, bottle, knife} and \emph{bowl}. Left: RGB image. Right: Estimated hand pose and configuration + estimated object pose and shape.}
  \label{fig-teaser}}
\makeatother

\author{Mia Kokic \and Danica Kragic \and Jeannette Bohg
\thanks{Mia Kokic and Danica Kragic are with the division of Robotics, Perception, and Learning, EECS, KTH, Stockholm, Sweden {\tt \{mkokic|dani\}@kth.se}}%
\thanks{Jeannette Bohg is with the Computer Science Department, Stanford University, CA, USA. {\tt bohg@stanford.edu}}%
\thanks{This work was partially supported by Wallenberg Autonomous Systems and Software Program (WASP), the Swedish Research Council and the Swedish Foundation for Strategic Research.}%
\thanks{Toyota Research Institute ("TRI") provided funds to assist the authors with their research but this article solely reflects the opinions and conclusions of its authors and not TRI or any other Toyota entity.}
}

\maketitle

\begin{abstract}
We develop a system for modeling hand-object interactions in 3D from RGB images that show a hand which is holding a novel object from a known category. 
We design a {\em Convolutional Neural Network\/} (CNN) for Hand-held Object Pose and Shape estimation called {\em HOPS-Net} and utilize prior work to estimate the hand pose and configuration. We leverage the insight that information about the hand facilitates object pose and shape estimation by incorporating the hand into both training and inference of the object pose and shape as well as the refinement of the estimated pose. The network is trained on a large synthetic dataset of objects in interaction with a human hand. To bridge the gap between real and synthetic images, we employ an image-to-image translation model ({\em Augmented CycleGAN}) that generates realistically textured objects given a synthetic rendering. 
This provides a scalable way of generating annotated data for training \emph{HOPS-Net}. 
Our quantitative experiments show that even noisy hand parameters significantly help object pose and shape estimation. 
The qualitative experiments show results of pose and shape estimation of objects held by a hand ``in the wild".



\end{abstract}


\section{Introduction}
\label{sec:introduction}
Observations of object poses during manipulation allow to infer contact interactions between objects and the hand. Contact interactions are central to understanding manipulation actions on a physical level. This level of understanding is crucial for applications such as object manipulation in virtual reality or when robots are learning to grasp objects from human demonstration \cite{kokic2017affordance, antonova2018global}.

While there exists a plethora of 2D images and videos of humans manipulating different objects~\cite{damen2018scaling, li2015delving,rogez2015understanding}, they often lack annotations that go beyond 2D bounding boxes and therefore do not allow a physical interpretation of manipulation actions. We aim to augment these datasets by lifting the data from 2D to 3D such that inferring physical properties of hand-object interactions becomes feasible. For this purpose, we propose an approach for estimating the shape and 6D pose of a hand-held object from an RGB image, see Fig.~\ref{fig-teaser}

This problem raises a number of challenges. First, the lack of depth data in the observations makes this problem highly under-constrained. Second, there are significant occlusions of the object and/or the hand during manipulation. Third, no real-world, large-scale dataset exists that is annotated with object and hand poses during manipulation actions. 
To alleviate some of these challenges we propose a system that: a) can operate on real images of novel objects instances from a known category while being trained only on synthetic data and b) leverages an estimate of hand pose and configuration~\cite{zimmermann2017learning} to facilitate estimation of the object pose and shape. 

We develop a CNN termed {\em HOPS-Net} which estimates the object pose and outputs shape features used to retrieve the most similar object from a set of 3D meshes. It leverages information about the hand in two ways: i) the hand parameters form an input to the network together with the segmented object to infer its pose and shape and ii) we use a model of the human hand to refine the initial object pose estimate and generate a plausible grasp in a grasping simulator. We train the network on images of synthetic meshes held by a hand in various poses. To address the \emph{sim-to-real gap}, we translate images of synthetic objects to multiple, realistic variants via image-to-image translation network. 
To our knowledge, this is the first application of such a network to address the \emph{sim-to-real gap} for objects which for the same shape can have many different appearances. 

Our contributions are as follows:
\begin{enumerate}[wide, labelwidth=!, labelindent=0pt]
    \item We develop a system for 3D shape and 6D pose estimation of a hand-held objects from a single RGB image. 
    \item Our approach can deal with novel object instances from four categories while being trained only on synthetic data. 
    \item We incorporate a human hand into training and refinement which we show improves the shape and pose estimates. This holds true even if the estimate of the hand pose and configuration are noisy. 
    \item We evaluate the accuracy of our method on our dataset of realistic objects held by a hand and demonstrate the applicability on a dataset of real world egocentric activities~\cite{rogez2015understanding}.
\end{enumerate}

\section{Related Work}
\label{subsec:relatedwork}

\subsection{Understanding Hand-Object Interactions}
Understanding hand-object interactions from vision is a long-standing problem that has been addressed from several angles. One line of work focuses on a hand alone, e.g., inferring hand pose and configuration from RGB or RGB-D data to either track the hand~\cite{romero2010hands, oikonomidis2011full, romero2013non} or infer manipulation actions~\cite{cai2015scalable, yang2015grasp, saran2015hand}. 
Another line of work follows the insight that the shape of an object influences the configuration of the hand and explores the hand in conjunction with the object~\cite{feix2014analysis, cai2016understanding, cai2017ego}.
Some authors even attempt to predict forces and contacts between the hand and the object~\cite{rogez2015understanding}.
However, scaling this approach to many objects would require large RGB or RGB-D datasets annotated with hand and object poses which is expensive to collect. 
Because of this, existing datasets contain either a small number of instances annotated with a 6D pose~\cite{garcia2017firstperson} or a large number of instances with coarse annotations such as grasp types or bounding boxes,~\cite{li2015delving, damen2018scaling} rendering them insufficient for in-depth understanding of hand-object interactions. In our work, we propose an approach for estimating both the hand and object pose from RGB which could potentially allow us to automatically annotate these datasets and use them for further analysis of hand-object interactions on a physical level.

\subsection{Object Shape and Pose Estimation}
To estimate object shape from RGB images, most recent works adopt a metric learning approach where the features of an image are matched against those of 3D mesh models rendered under multiple viewpoints. Shape estimation is based on finding the nearest neighbor among features of the rendered meshes and retrieving it~\cite{oh2016deep, aubry2015understanding, izadinia2017CVPR}.
Another approach is to map 3D voxel grids and an image into one low-dimensional embedding space. 3D model retrieval is then performed by embedding a real RGB image and finding the nearest neighbor among voxel grids~\cite{girdhar2016learning}. We adopt this approach since it allows us to use all the available information about the shape instead of just renderings from different viewpoints. We embed the 2D and 3D instances in the same latent space and force the 2D and 3D features of the same object to be close in this space and distant otherwise.

Research on 6D pose estimation of objects can be grouped based on two aspects: i) input modality (2D, 2.5D, 3D, multi-modal) and ii) prior object knowledge (known, unknown).
When the object is known and the input is an RGB image, the main challenge is to resolve ambiguities that stem from object symmetries as well as the orientation representations (e.g., Euler angles or quaternions). 
State-of-the-art methods use CNNs for directly regressing position and orientation~\cite{xiang2017posecnn, kehl2017ssd, brachmann2016uncertainty} or implicitly learning pose descriptors~\cite{sundermeyer2018implicit, balntas2017pose}. 
The latter has an advantage over direct regression since it can handle ambiguous representations and poses caused by symmetric views. The initial predictions are often refined using {\em Iterative Closest Point\/} (ICP) if depth data is available~\cite{xiang2017posecnn, brachmann2016uncertainty,sundermeyer2018implicit}.
In our work, we directly regress the position and orientation from RGB images but resolve ambiguities by constructing a continuous, category-specific representation of the object orientation that is invariant to the symmetries of this specific object category.

When the object is new but belongs to a known category, the biggest challenge is to develop a method that can cope with large intra-class variations and can generalize to novel instances. 
To do this, the datasets for training have to contain many 3D shapes annotated with pose labels. This is often done by manually aligning meshes of objects with objects in RGB images~\cite{tulsiani2015pose,grabner20183d} which is time-consuming and expensive. 
A way to address this is to train the model on synthetic data, e.g., on the renderings of meshes in various poses. Pose annotations of the resulting images are for free. The caveat is that the synthetic images is so different in appearance from real images that models trained on it typically do not generalize well to reality. 

\subsection{Bridging the Sim-to-Real Gap} 
There are several strategies for making synthetic data more realistic. For example, photo-realistic rendering and domain randomization techniques~\cite{kehl2017ssd, hinterstoisser2018pre, tremblay2018deep} augment synthetic images with different backgrounds, random lighting conditions, etc.  
This is often sufficient to reduce the discrepancy if the gap between synthetic and real data is already relatively small, e.g., when synthetic objects are already textured.
In our work, we circumvent the problem of manually annotating the real data to train the model. We use only images of synthetic objects for which the annotations are automatically available but translate it into realistic looking images with a variant of a {\em Generative Adversarial Network.}

\section{Problem Statement}
\begin{figure*}[h!]
\centering
\includegraphics[width=\textwidth]{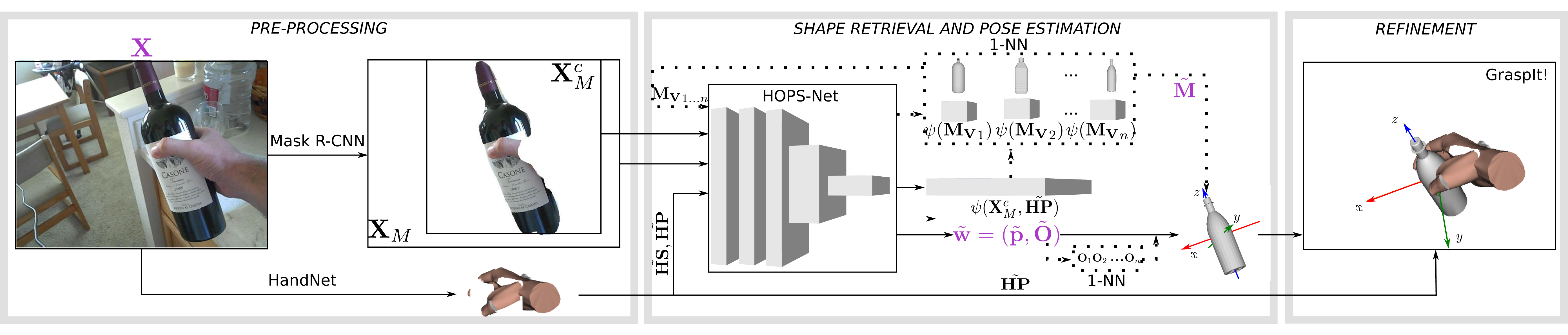}
\caption{\label{fig:system} The input to the system is an image $\mathbf{X}$ showing a hand holding an object and the output is the object shape $\tilde{\mathbf{M}}$ and the pose $\tilde{\mathbf{w}}$ (color purple). In the pre-processing stage we use existing methods to obtain the object category $c$, a segment of the object $\mathbf{X}_M$ and an estimate of the hand shape $\tilde{\mathbf{HS}}$ as well as hand pose $\tilde{\mathbf{HP}}$. In the second stage, our model outputs the object position $\tilde{\mathbf{p}}$, orientation $\tilde{\mathbf{O}}$ and a feature vector $\psi(\mathbf{X}_{M}^{c}, \tilde{\mathbf{HP}})$ which is used to retrieve the most similar mesh in the training data examples (dotted lines). We also find the closest orthogonal rotation matrix in the training examples (retrievals are denoted with dotted rectangles). Finally, we refine the initial object pose such that the fingertips of the hand are in contact with the object surface and compute a plausible grasp.}
\end{figure*}

Our method estimates the 3D shape and 6D pose of a novel hand-held object instance from a single RGB image. We consider four categories of objects: \emph{bottle}, \emph{mug}, \emph{knife} and \emph{bowl}. We follow the insight that information about the hand helps estimation of object pose and shape. Our category-specific CNN \emph{HOPS-Net} takes as input (i) an RGB image of a segmented object and (ii) the estimated hand pose and configuration. It outputs the position and orientation of the object in the camera frame as well as shape features which we use to retrieve the most similar looking mesh from the training data. 

\subsection{Notation}
\subsubsection{Image} Let $\mathbf{X} \in \mathbb{R}^{H \times W \times 3}$ be an RGB image of an object held by a hand. We denote the image showing the segmented object on a white background with $\mathbf{X}_{M}$ and the minimum square bounding box around the segment, i.e. the crop, with $\mathbf{X}_{M}^{c}$.

\subsubsection{Hand} Let $\mathbf{HP} \in \mathbb{R}^{3} \times SE(3) \times \mathbb{R}^{20}$ describe the human hand pose in the camera frame and its joint configuration. The pose is described by wrist position and orientation in form of a rotation matrix. The configuration is described by $20$ joint angles. Let $\mathbf{HS} \in \mathbb{R}^{20 \times 3}$ describe the hand shape in that joint configuration. $\mathbf{HS}$ contains the 3D position of each of the $20$ joints in the camera frame. 

\subsubsection{Object} Let $\mathbf{M}$ be a polygonal mesh and $\mathbf{M}_{V} \in \mathbb{Z}_{2}^{H \times W \times D}$ its voxelized representation i.e., the binary occupancy grid. Let $\mathbf{w} = (\mathbf{p}, \mathbf{O}) \in SE(3)$ be the pose of the object, i.e. its position $\mathbf{p} \in \mathbb{R}^3$ and the orientation $\mathbf{O} \in SO(3)$\footnote{For definitions of SE(3) and SO(3) see \cite{cederberg2013course}, \cite{jacobson2009basic}} in the camera frame. We use $\mathbf{R}$ to denote a category-specific orientation representation which consists of the columns in $\mathbf{O}$.

Features are denoted by $\psi(\cdot)$, estimated variables by $^\sim$.

\subsection{System Overview}
The input to the system is an RGB image $\mathbf{X}$ of a hand holding a novel object. The system estimates the object shape $\tilde{\mathbf{M}}$ and the pose $\tilde{\mathbf{w}}$. Fig.~\ref{fig:system} illustrates our complete model. 
Our approach consists of three stages. In the first, pre-processing stage, we leverage existing approaches for hand estimation and instance segmentation. We use a network for hand pose and configuration estimation~\cite{zimmermann2017learning} which we call \emph{HandNet} to estimate hand shape $\tilde{\mathbf{HS}}$ as well as pose and configuration $\tilde{\mathbf{HP}}$. We use \emph{Mask R-CNN}~\cite{he2017mask} to obtain the object category $c$ and a segment of the object $\mathbf{X}_M$. In the second stage, our model outputs the object position $\tilde{\mathbf{p}}$, orientation $\tilde{\mathbf{O}}$ and a shape feature vector $\psi(\mathbf{X}_{M}^{c}, \tilde{\mathbf{HP}})$ which is used to retrieve the most similar mesh in the training data examples. In the third stage, we optimize the initial object pose such that the fingertips of the estimated hand are in contact with the object surface to obtain a plausible grasp.

\section{Pose Estimation and Shape Retrieval}
\label{subsec:pse}
To estimate object shape $\tilde{\mathbf{M}}$ and pose $\tilde{\mathbf{w}}$ from an image $\mathbf{X}$, we develop a single model that learns three mappings: 
\begin{align}
    f: &(\mathbf{X}_{M}^{c}, \tilde{\mathbf{HP}}) \mapsto  \psi(\mathbf{X}_{M}^{c}, \tilde{\mathbf{HP}}) \label{eq:shape}\\
    g: &(\mathbf{X}_{M}, \tilde{\mathbf{HS}}) \mapsto  \tilde{\mathbf{p}} \label{eq:position}\\
    h: &(\mathbf{X}_{M}^{c}, \tilde{\mathbf{HP}}) \mapsto \tilde{\mathbf{O}} \label{eq:orientation}
\end{align}

The estimation problem benefits from the information about the hand in several ways. First, the hand provides cues about object position even when the object is occluded by the hand. 
Second, hand orientation and shape provide cues about object orientation and shape, e.g., a knife is often grasped at the handle such that the blade is pointing downwards. Therefore, information on the hand forms the input to all three mappings. To estimate the position, Eq.~(\ref{eq:position}) takes as input the segmented object $\mathbf{X}_M$ and hand shape $\tilde{\mathbf{HS}}$. To estimate object shape and orientation, Eq.~(\ref{eq:shape}) and Eq.~(\ref{eq:orientation}) take as input the cropped object segment $\mathbf{X}_{M}^{c}$ as well as hand pose and joint configuration $\tilde{\mathbf{HP}}$.
We found empirically, that the 3D joint positions are more indicative of object position while the hand pose and joint configuration help determine the object shape and orientation\footnote{Assuming that the hand is wrapped around the object, the 3D positions of the hand joints are indicative of the object centroid. Furthermore, object orientation can be expressed as a transformation of the hand orientation matrix.}.

The model is trained with the objective of minimizing the sum of the shape, position and orientation losses.
\begin{equation}
    \mathcal{L} = \mathcal{L}_{\mathbf{M}} +  \mathcal{L}_{\mathbf{p}} + \mathcal{L}_{\mathbf{O}}
\end{equation}

\subsection{From a 2D image to 3D Shape}
For estimating object shape, we adopt a metric learning approach that generates a joint embedding space for shape and image features. Our model is based on a {\em Siamese network} that is used for dimensionality reduction in weakly supervised metric learning~\cite{koch2015siamese}. It consists of two twin networks that have the same structure and share weights. 

For training, the network takes two data points and outputs their features which are used to compute the loss. The loss ensures that similar data points are mapped close to each other in the embedding space and distant otherwise. In our model, the first data point is a crop of the segmented object $\mathbf{X}_{M}^{c}$ as well as the hand pose and its configuration $\mathbf{HP}$. The second data point is a random mesh $\mathbf{M_V}$. Each data point is pre-processed to obtain features of the same dimension, and a {\em Siamese network} network outputs $\psi(\mathbf{X}_{M}^{c}, \mathbf{HP})$ and $\psi(\mathbf{M_V})$, which are used to compute the shape loss $\mathcal{L}_{\mathbf{M}}$. This loss is defined as follows:
\begin{equation}
    \mathcal{L}_{\mathbf{M}} = (1 - y)\frac{1}{2}d^2 + \frac{1}{2}{max(0, m - d)}^2,
\end{equation}
where $d$ is a Euclidean distance between the features $\psi(\mathbf{X}_{M}^{c}, \mathbf{HP})$ and $\psi(\mathbf{M_V})$, $m$ is a margin as defined in~\cite{hadsell2006dimensionality} and $y$ is a binary valued function which is equal to $1$ if $\mathbf{X}_{M}^{c}$ and $\mathbf{M_V}$ belong to the same object, and $0$ otherwise. 

At test time, we only have access to the crop of the segmented object and the estimated hand to compute $\psi(\mathbf{X}_{M}^{c}, \tilde{\mathbf{HP}})$. We obtain an estimate of the 3D object shape $\tilde{\mathbf{M_V}}$ by comparing $\psi(\mathbf{X}_{M}^{c}, \tilde{\mathbf{HP}})$ to the pre-computed $\psi(\mathbf{M_V})$ of all the meshes in the training dataset. $\tilde{\mathbf{M}}$ will be the nearest neighbour in that feature space.

\subsection{From a 2D Image to 6D Pose}
We estimate the pose of an object by directly regressing the $\tilde{\mathbf{p}}$ and $\tilde{\mathbf{O}}$. Since these two variables are concerned with different visual properties, i.e., the location of the object in the image vs. features that are informative of the orientation, we learn them separately. 

\begin{figure}[h]
\centering
\scalebox{0.65}{
\begin{tabular}{c|c|c}
$c$ & \multicolumn{1}{c|}{\begin{tabular}[c]{@{}c@{}}Symmetry\end{tabular}} & \multicolumn{1}{c}{\begin{tabular}[c]{@{}c@{}} $\mathbf{R}$\end{tabular}} \\ \hline
\includegraphics[align=c, width=0.1\textwidth]{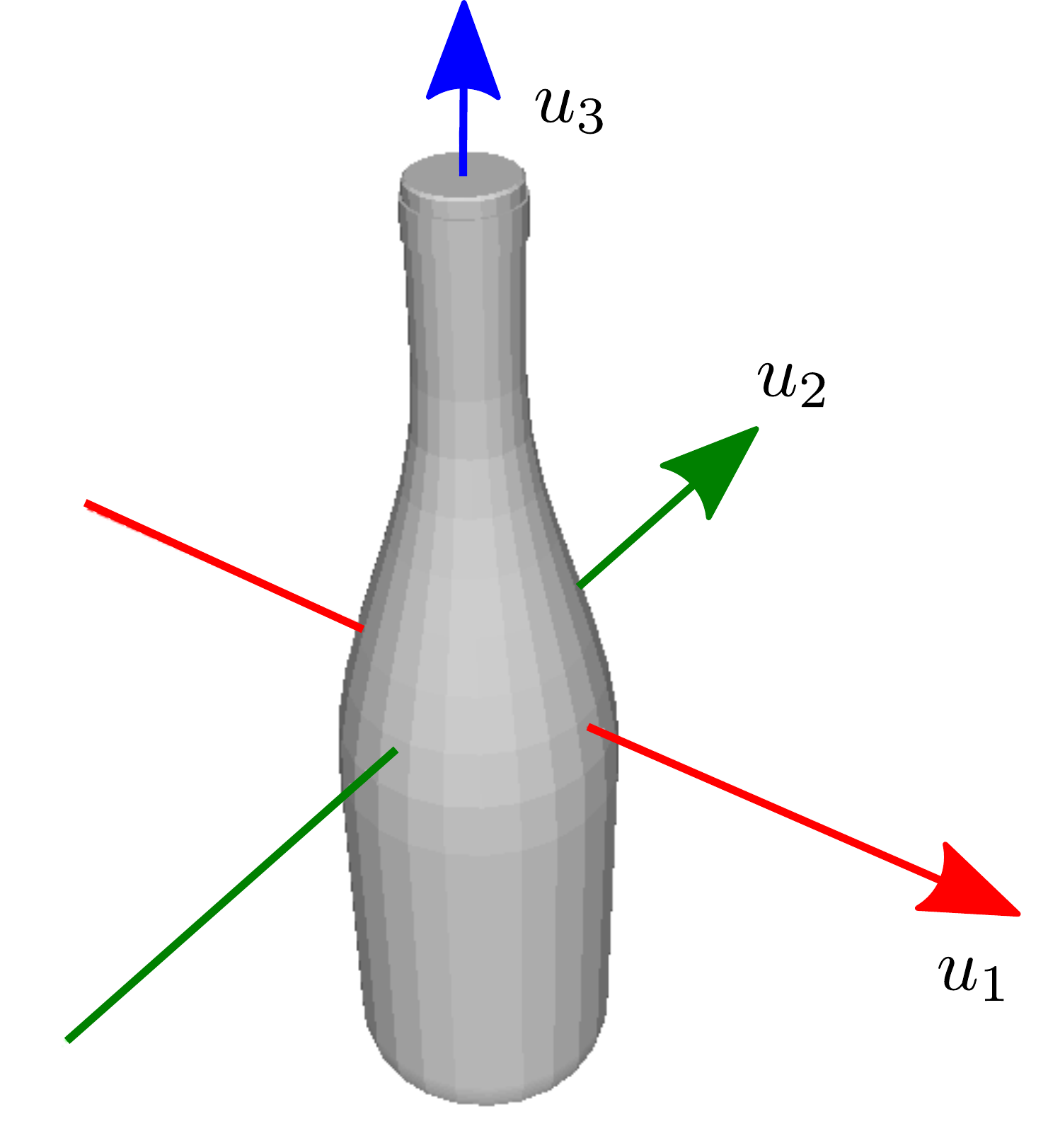} & axial spherical & $\mathbf{R} = \mathbf{u_3}$ \\
\includegraphics[align=c, width=0.1\textwidth]{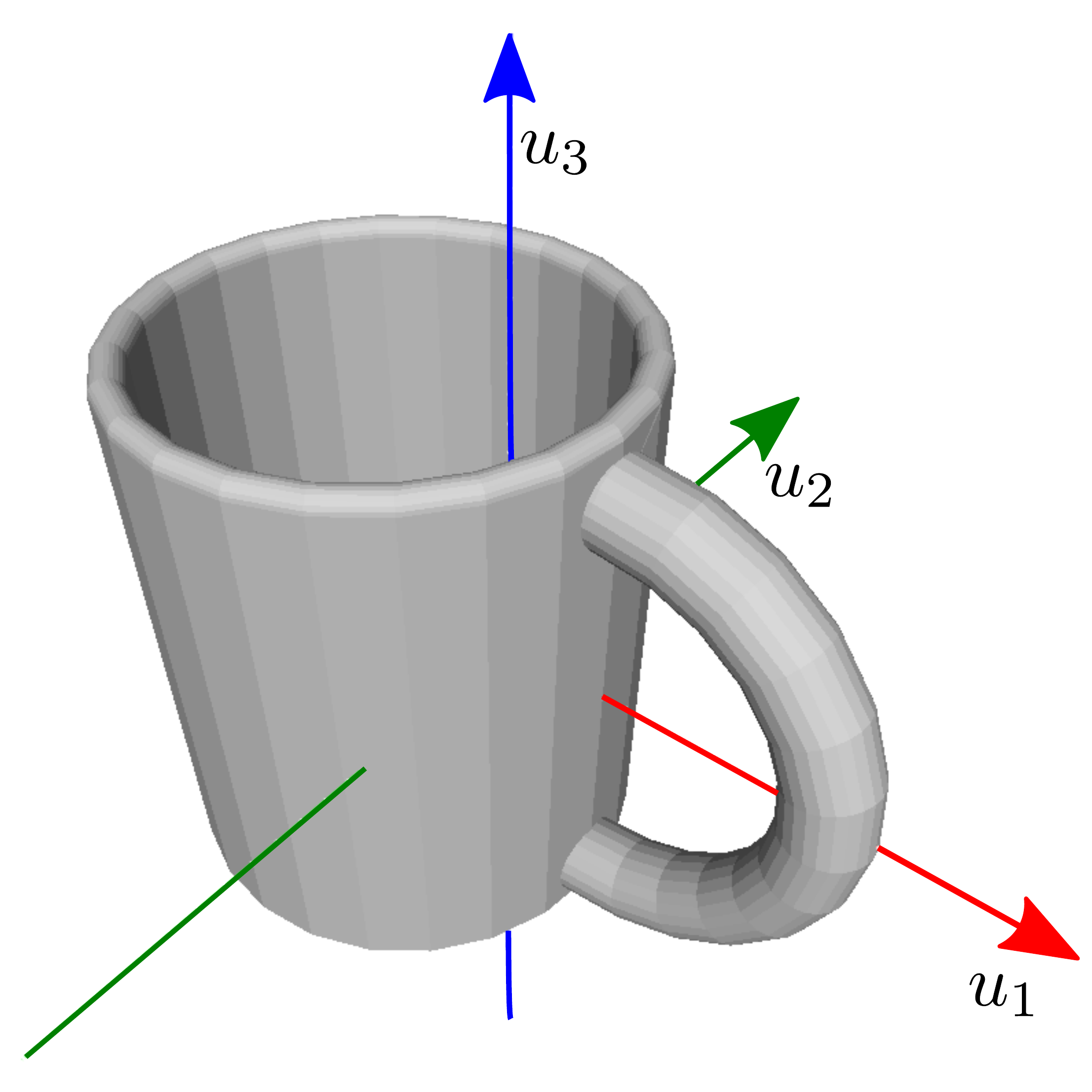} & single plane reflection & $\mathbf{R} = ([\mathbf{u_1}; \mathbf{u_3}], \mathbf{u_2} \mathbf{u_2}^T)$ \\ 
\includegraphics[align=c, width=0.1\textwidth]{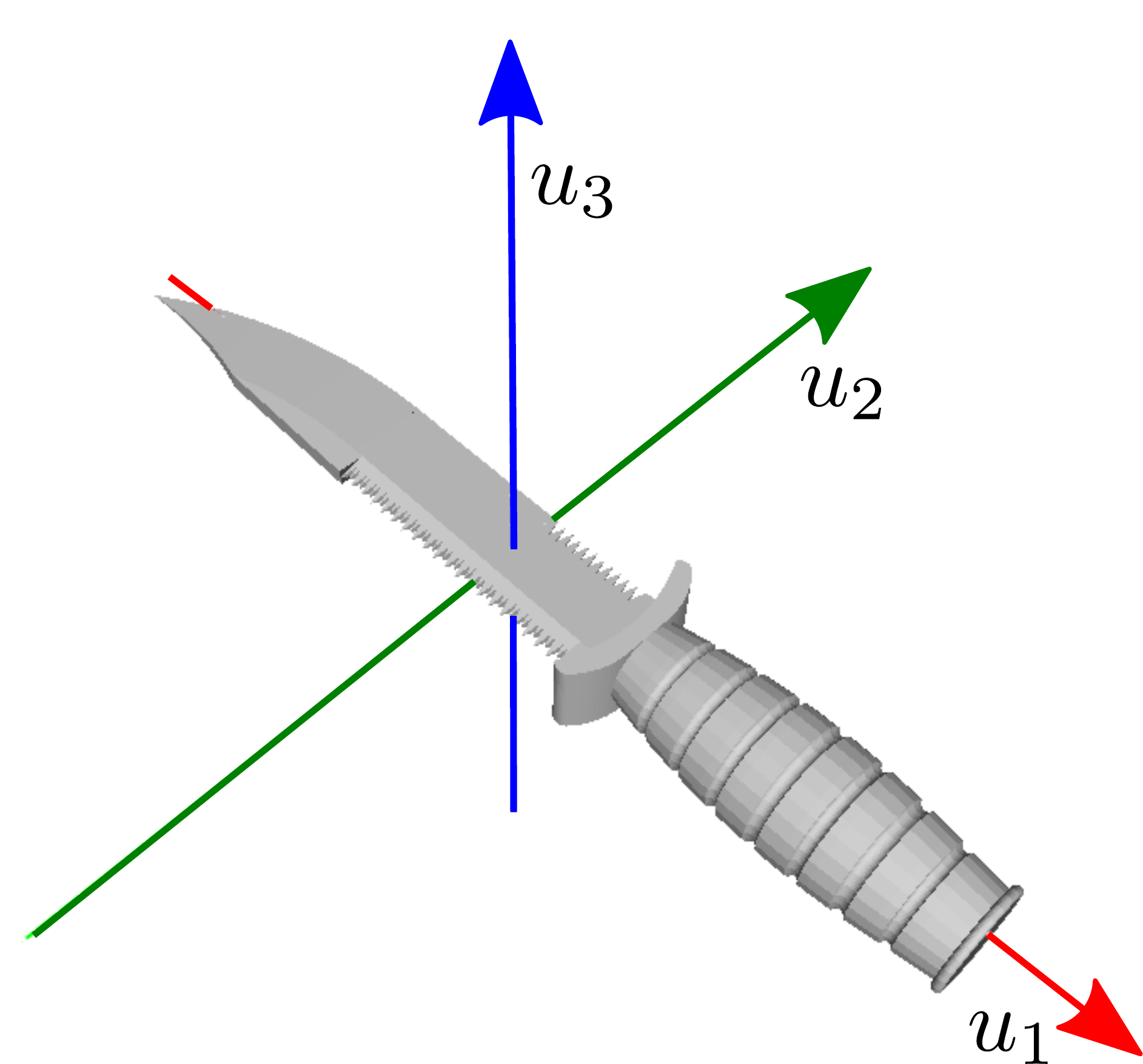} & single plane reflection & $\mathbf{R} = ([\mathbf{u_1}; \mathbf{u_2}], \mathbf{u_3} \mathbf{u_3}^T)$ \\ 
\includegraphics[align=c, width=0.1\textwidth]{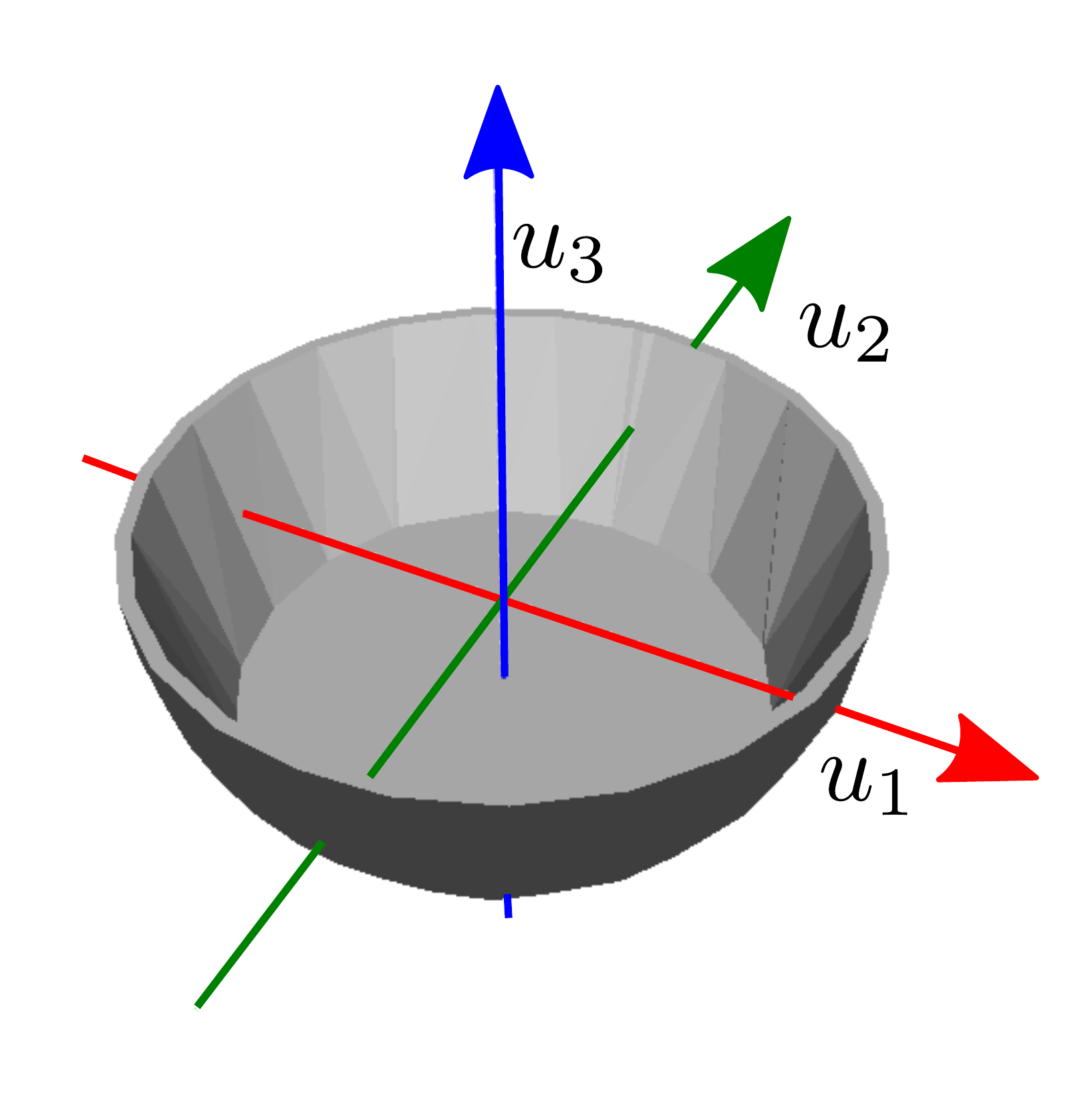} & axial spherical & $\mathbf{R} = \mathbf{u_3}$ \\ 
\end{tabular}
}
\caption{Per category orientation representation $\mathbf{R}$. It consists of columns $\mathbf{u}$ of a rotation matrix $\mathbf{O}$ and is constructed based on the type of symmetry that the object category possesses.}
\label{fig:symmetry}
\end{figure}

As discussed in Sec.~\ref{subsec:relatedwork}, computing the orientation loss over the whole rotation matrix is challenging due to ambiguities in the representation and object symmetry. However, with a representation that is invariant to symmetry we can facilitate the learning process. Following the work of~\cite{saxena2009learning}, we construct a category-specific representation $\mathbf{R}$ which consists of columns $\mathbf{u}$ of a rotation matrix and is defined depending on the type of symmetry that an object possesses, see Fig.~\ref{fig:symmetry}. 

For each category, we compute the orientation loss by extracting the columns $\tilde{\mathbf{u}}$ from the estimated orientation $\tilde{\mathbf{O}}$ and constructing $\tilde{\mathbf{R}}$. The loss $\mathcal{L}_{\mathbf{O}}$ is the mean squared error between $\tilde{\mathbf{R}}$ and $\mathbf{R}$. Therefore, estimated orientations $\tilde{\mathbf{O}}$ that differ from the ground truth orientation $\mathbf{O}$ are not penalized if they yield the same $\mathbf{R}$. The position loss $\mathcal{L}_{\mathbf{p}}$ is also computed as mean squared error between $\tilde{\mathbf{p}}$ and $\mathbf{p}$.

\vspace{-0.5cm}
\section{Data Generation}
\label{sec:datagen}
To train \emph{HOPS-Net}, we need realistic RGB images of objects that are annotated with poses and shapes as well as information on the hand. In Sec.~\ref{subsec:augcgan} we describe how we generate realistic images from textureless meshes and in Sec.~\ref{subsec:annot} how we generate pose and shape labels.

\subsection{Generating Realistic Images}
\label{subsec:augcgan}
A naive way to generate the data for training would be to render projections of grasped 3D meshes in various poses. However, training the network on synthetic images would not generalize well to real images due to the large visual discrepancy them. This discrepancy can be reduced by translating the synthetic images to real images and annotating them. 
To do this, we use a variant of \emph{CycleGAN} - an image-to-image translation network~\cite{zhu2017unpaired} which maps images from the source domain $Y$ to the target domain $X$ and vice versa. 
The crucial benefit of CycleGAN is that it can learn from unpaired data i.e. without explicit information on which data point from $Y$ matches a data point from $X$. To constrain this problem, it uses a cycle consistency loss which ensures that the forward and backward mappings are bijective. 

\begin{figure}[t]
\centering
\includegraphics[width=0.4\textwidth]{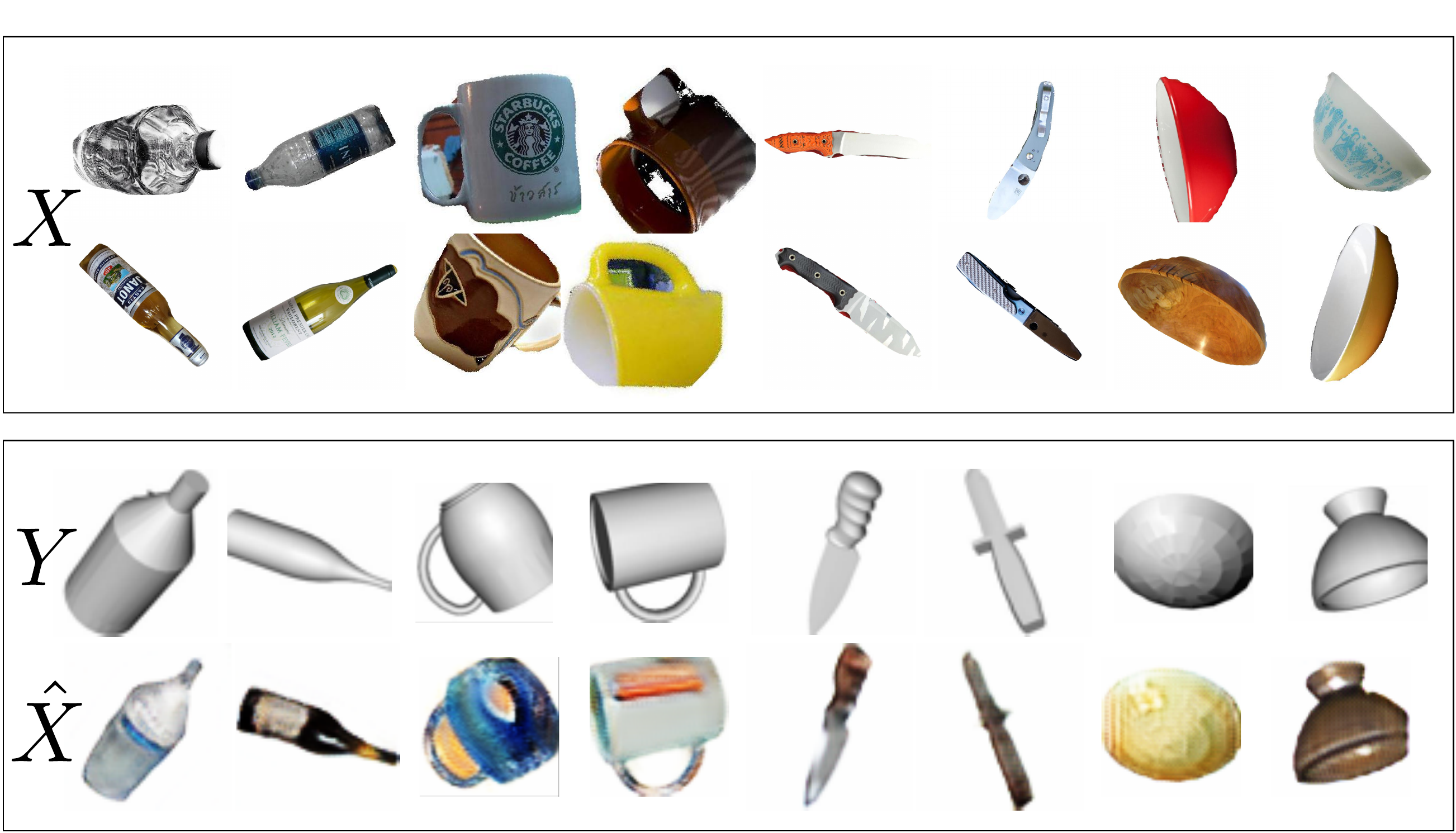}
\caption{Top: Example images in the real domain. Bottom: Top row shows example images in the synthetic domain (two per category) and the bottom row shows generated real world variants. The generated images look realistic both in color and texture, e.g, a label on a bottle.}
\label{fig:cgan}
\end{figure}

\emph{CycleGAN} can only handle one-to-one mappings which is unsuitable for our scenario where an image of a synthetic object can be mapped to many real variants, e.g., an image of a synthetic bottle can be mapped to many real bottles of different colors and textures. 
Therefore, we propose to use an augmented \emph{CycleGAN} which can produce one-to-many and many-to-many mappings between domains. It learns these mappings by augmenting $Y$ and $X$ with latent spaces $Z_x$ and $Z_y$ respectively. The latent spaces have a Gaussian prior over their elements and as such can capture variations in both domains (see Fig.~\ref{fig:cgan}). 
We constrain the model to learn one-to-many mappings between synthetic and realistic images by learning a function $i:Y \times Z_x \mapsto X$. Cycle consistency is achieved with two losses, one for recovering $Y$ and one for recovering $Z_x$. For details see~\cite{almahairi2018augmented}.

To train the \emph{AugCGAN}, for each category we construct the source domain $Y$ by projecting the gray, textureless 3D meshes in various poses onto a 2D image. The target domain $X$ is a collection of images from ImageNet~\cite{deng2009imagenet} that are pre-processed with \emph{Mask R-CNN} to generate object segments. In both domains objects are cropped and randomly rotated. The \emph{AugCGAN} outputs ``real" images of synthetic objects that are used to generate a dataset for \emph{HOPS-Net}.  
We denote an instance of the source domain with $\mathbf{Y}$, the target domain with $\mathbf{X}$ and the translated ``real" image with $\hat{\mathbf{X}}$. The resulting model allows us to generate realistic training data for \emph{HOPS-Net} without requiring laborious, manual annotations. 

\subsection{Generating Annotations for HOPS - Net}
\label{subsec:annot}
We use the \emph{GraspIt!}~\cite{miller2004graspit} simulation environment to generate a dataset of hand-held objects. Thereby, the generated images $\hat{\mathbf{X}}_{M}$ and $\hat{\mathbf{X}}_{M}^{c}$ show the segmented object often severely occluded by the hand. Each image is annotated with hand configurations $\{\mathbf{HS}$, $\mathbf{HP}\}$ as well as object pose and shape $\{\mathbf{p}, \mathbf{O}, \mathbf{M}\}$. We refer to the dataset generated in this manner as the ``real" occluded dataset.

To generate one data point, we import a mesh $\mathbf{M}$ into \emph{GraspIt!} and render the object in pose $\mathbf{w}$ to yield the image $\mathbf{Y}_{M}$. We then crop the image and run it through the \emph{AugCGAN} to obtain $\hat{\mathbf{X}}_{M}^{c}$.
We also save the coordinates of the crop center so that we can generate $\hat{\mathbf{X}}_{M}$ by positioning $\hat{\mathbf{X}}_{M}^{c}$ into the original image. 
We then import the hand parameterized with $\mathbf{HP}$ which is randomly sampled from a set $\mathcal{HP}$ containing configurations of stable grasps. The grasps are generated off-line with the EigenGrasp planner~\cite{EigenGrasp}. 
We manually constrain the set by selecting grasps that satisfy two conditions: i) all fingertips are in contact with the object surface and ii) generated grasps follow a real world distribution of task-specific grasps for a certain category, e.g., mugs are grasped either from the top or from the side with the thumb and index finger close to the opening since these grasps allow \emph{lifting} or \emph{pouring}. 

When testing on real world images, we can expect noisy hand estimates. To better account for that, we augment the dataset by adding Gaussian noise to all parameters in $\mathbf{HP}$ and render an image of the resulting hand-object configuration. We then mask out the hand by computing the pixel-wise difference between the image with and without the hand. All pixels showing a difference are coloured white. This yields $\hat{\mathbf{X}}_{M}$. We also crop this image to obtain $\hat{\mathbf{X}}_{M}^{c}$. If the hand occludes less than $50\%$ of the object, we save the images as a training data point and compute $\mathbf{HS}$ from $\mathbf{HP}$ through forward kinematics of the human hand model. Otherwise, we proceed with the next instance. 

\section{Training and Inference}
\label{subsec:training}
\subsection{Network Architecture}
\begin{figure}
\centering
\scalebox{0.8}{
\includegraphics[width=0.5\textwidth]{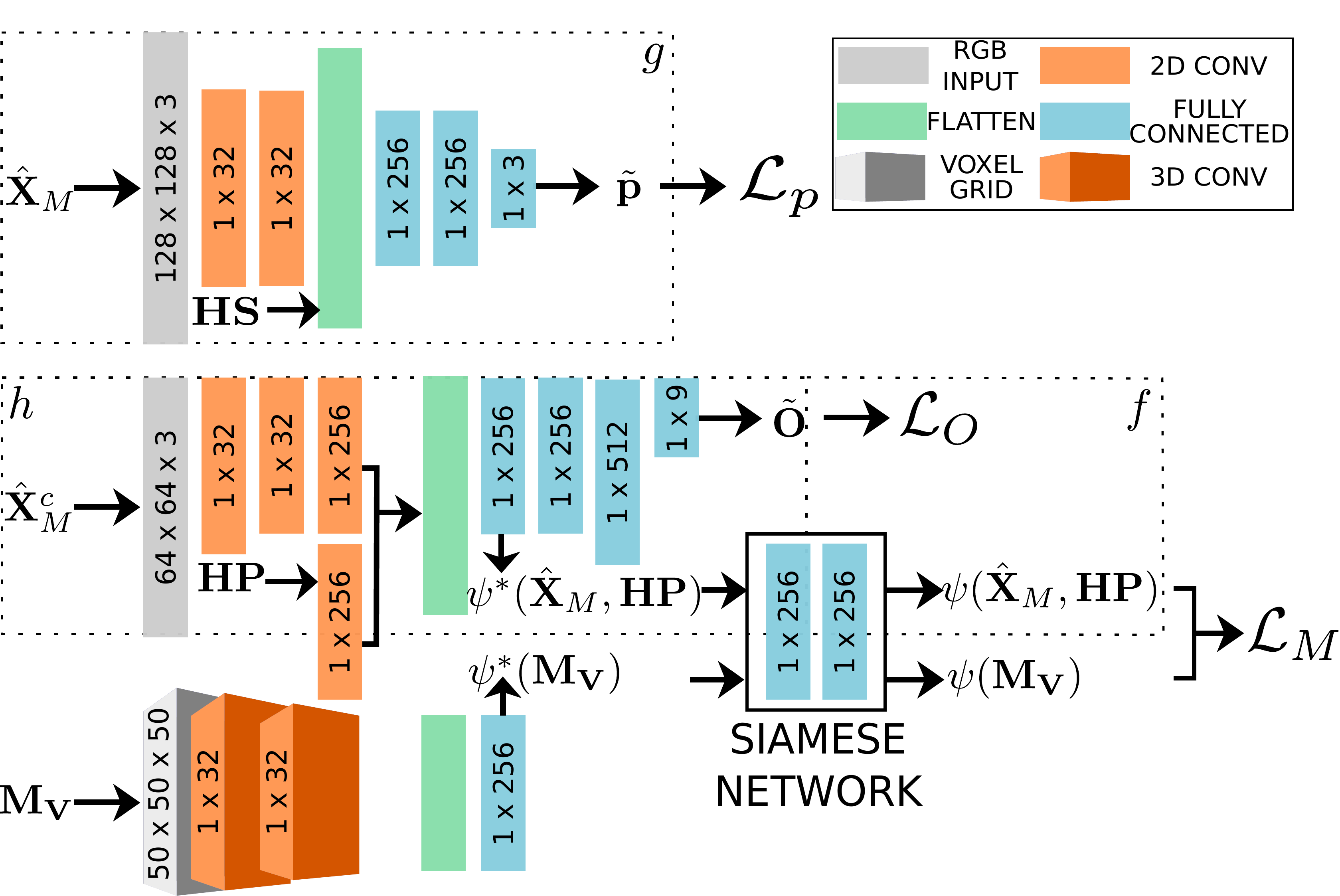}
}
\caption{\emph{HOPS-Net} architecture. Dotted rectangles show mappings $f: (\hat{\mathbf{X}}_{M}^{c}, \mathbf{HP}) \mapsto  \psi(\hat{\mathbf{X}}_{M}^{c}, \mathbf{HP})$, $g: (\hat{\mathbf{X}}_{M}, \mathbf{HS}) \mapsto  \tilde{\mathbf{p}}$ and $h: (\hat{\mathbf{X}}_{M}^{c}, \mathbf{HP}) \mapsto \tilde{\mathbf{O}}$.}
\label{fig:netarch}
\end{figure}

During training, {\em HOPS-Net\/} gets as input an object segment $\hat{\mathbf{X}}_{M}$, its crop $\hat{\mathbf{X}}_{M}^{c}$ and a binary voxel grid $\mathbf{M}_{\mathbf{V}}$ of a random mesh in its nominal orientation. We run $\hat{\mathbf{X}}_{M}$ and $\hat{\mathbf{X}}_{M}^{c}$ through a set of independent convolutional layers, flatten the outputs and append them with $\mathbf{HS}$ and $\mathbf{HP}$ respectively. 
$\hat{\mathbf{X}}_{M}$ and $\mathbf{HS}$ are passed through two three connected layers to estimate the object position $\tilde{\mathbf{p}}$. $\hat{\mathbf{X}}_{M}^{c}$ and $\mathbf{HP}$ are also passed through a fully connected layer to extract intermediate features $\psi^{*}(\hat{\mathbf{X}}_{M}^{c}, \mathbf{HP})$ and then another three to get the estimate of the orientation $\tilde{\mathbf{O}}$. 
Simultaneously, we run $\mathbf{M}_{\mathbf{V}}$ through two 3D convolutional layers, flatten the output, and run it through a fully connected layer to get the intermediate feature vector $\psi^{*}(\mathbf{M}_{\mathbf{V}})$ that has the same dimension as $\psi^{*}(\hat{\mathbf{X}}_{M}^{c}, \mathbf{HP})$. Intermediate features are then passed through the {\em Siamese network} and the final image $\psi(\hat{\mathbf{X}}_{M}^{c}, \mathbf{HP})$ and shape features $\psi(\mathbf{M}_{\mathbf{V}})$ are extracted, see Fig.~\ref{fig:netarch}.

\subsection{Inference}
For inference, $\mathbf{X}$ is first pre-processed with \emph{Mask R-CNNN} to obtain $\mathbf{X}_{M}$,  $\mathbf{X}_{M}^{c}$ and the object category $c$. We use \emph{HandNet} to estimate the normalized 3D coordinates of the joints and the position of the wrist. We scale these values to comply with the hand model in \emph{GraspIt!}, generate $\tilde{\mathbf{HS}}$ and compute the hand pose and configuration $\tilde{\mathbf{HP}}$ through inverse kinematics. 
We assume that \emph{Mask R-CNN} can segment the object from the background with high accuracy and that the values obtained with \emph{HandNet} are close to the ground truth values. 

\emph{HOPS-Net} estimates $\tilde{\mathbf{p}}$, $\tilde{\mathbf{O}}$ and outputs $\psi(\mathbf{X}_{M}^{c}, \tilde{\mathbf{HP}})$. 
To retrieve the best matching object shape, we minimize the Euclidean distance between $\psi(\mathbf{X}_{M}^{c}, \tilde{\mathbf{HP}})$ and the pre-computed $\psi(\mathbf{M_V})$ for all meshes in the training dataset. Furthermore, since $\tilde{\mathbf{O}}$ is not orthogonal we find the nearest orthogonal matrix by also finding the closest rotation matrix from the training set\footnote{We also tested matrix orthogonalization~\cite{saxena2009learning} but the final orientation estimates were better with the nearest neighbor approach presumably because the estimated rotation matrices are far from being orthogonal.}.

\subsection{Pose Refinement}
\label{subsubsec:refinement}
Once we have estimated the object pose, we refine it by regressing it into the hand and thereby to generate a plausible grasp that can be analyzed in GraspIt!. To do this, we import the human hand model and the retrieved object into the simulator in $\tilde{\mathbf{w}}$ and $\tilde{\mathbf{HP}}$ respectively. We assume that the hand pose is fixed and optimize the object pose by minimizing the distance between fingertip locations and points on the object surface.
\begin{equation}
    \operatorname*{arg\,min}_{\tilde{\mathbf{O}}, \tilde{\mathbf{p}}} \frac{1}{5} \sum_{\mathbf{x}_f \in \mathbf{HC}} \operatorname*{min}_{\mathbf{x}_m \in \mathbf{M}}||\mathbf{x}_f - (\tilde{\mathbf{O}}\mathbf{x}_m + \tilde{\mathbf{p}})||
\end{equation}
Here, $\mathbf{x}_f \in \mathbb{R}^3$ denotes the fingertips positions and $\mathbf{x}_m \in \mathbb{R}^3$ the positions of mesh vertices. The optimization process is constrained to $5$cm in position and $5\deg$ in orientation of the object. This yields a refined pose of the object but still does not guarantee that the hand and object are not intersecting. To address this, we then keep the object pose fixed run the planner that computes a physically plausible, force-closure grasp. 

\section{Evaluation}
\label{sec:evaluation}
We quantitatively evaluate the accuracy of shape retrieval and pose estimation on the ``real" occluded dataset. To evaluate how much information on the hand helps object pose estimation, we perform an ablation study for training the model with and without the hand. We show qualitative results on a subset of images ``in the wild" from a dataset of manipulation actions~\cite{rogez2015understanding}. 

\begin{figure}
\centering
\includegraphics[width=0.5\textwidth]{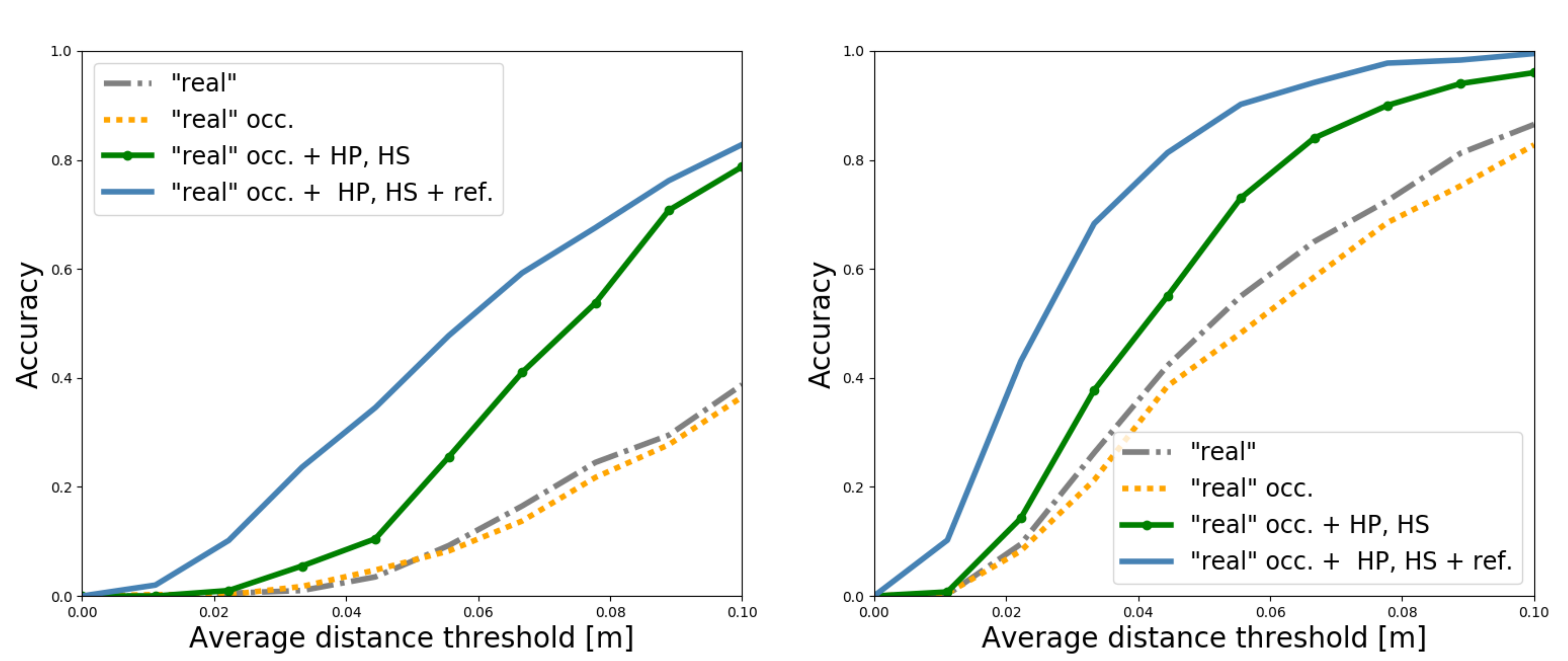}
\caption{Accuracy-threshold curves for accuracy computed with the ADD (left) and ADD-s (right) metric. The curves are plotted for training cases i) - iv).}
\label{fig:curves}
\end{figure}

\begin{figure}[h]
\centering
\scalebox{0.7}{
\begin{tabular}{ccccccc}
\multicolumn{1}{l}{} & \multicolumn{3}{c}{ADD} & \multicolumn{3}{c}{ADD-S} \\ \hline
$c$ & 
\multicolumn{1}{c}{\begin{tabular}[c]{@{}c@{}}``real" \\occ.\end{tabular}} &
\multicolumn{1}{c}{\begin{tabular}[c]{@{}c@{}}``real" \\occ. \\+ \\$\mathbf{HP, HS}$\end{tabular}} & \multicolumn{1}{c}{\begin{tabular}[c]{@{}c@{}}``real" \\occ. \\+ \\$\mathbf{HP, HS}$ ref.\end{tabular}} &
\multicolumn{1}{c}{\begin{tabular}[c]{@{}c@{}}``real" \\occ.\end{tabular}} &
\multicolumn{1}{c}{\begin{tabular}[c]{@{}c@{}}``real" \\occ. \\+ \\$\mathbf{HP, HS}$\end{tabular}} & \multicolumn{1}{c}{\begin{tabular}[c]{@{}c@{}}``real" \\occ. \\+ \\$\mathbf{HP, HS}$ ref.\end{tabular}} \\ \hline
bottle & 13.51 & 39.25 & 53.88 & 34.69 & 64.13 & 75.67\\ 
mug & 8.47 & 21.33 & 23.38 & 53.45 & 67.77 & 71.77\\ 
knife & 7.05 & 30.11 & 42.27 & 28.73 & 55.94 & 73.61\\  
bowl & 16.77 & 21.00 & 31.44 & 52.38 & 56.44 & 66.88 \\  \hline
\textbf{avg} & 11.45 & 27.92 & 37.44 & 42.31 & 61.07 & 71.98
\end{tabular}
}
\caption{Area under the accuracy-threshold curve for ADD and ADD-s metrics. The results correspond to training cases ii)-iv).}
\label{fig:abl}
\end{figure}

\subsection{Metrics}
To evaluate the accuracy of object pose estimation, we first compute the average pairwise distance between the mesh surface points in the ground truth and estimated pose (also known as average distance (ADD) metric~\cite{hinterstoisser2012model}),
\begin{equation}
    \text{ADD} = \frac{1}{m} \sum_{\mathbf{x}_m \in \mathbf{M}} ||(\mathbf{O}\mathbf{x}_m + \mathbf{p}) - (\tilde{\mathbf{O}}\mathbf{x}_m + \tilde{\mathbf{p}})||,
\end{equation}
where $\mathbf{x}_m \in \mathbf{M}$ are mesh surface points.
A pose estimate is considered correct if this average is below a certain threshold. We vary the threshold from $0$ to $10$cm as in~\cite{xiang2017posecnn} and for each value, determine the number of correct examples in the test set. From this, we can compute the area under the accuracy-threshold curve.
A problem with the ADD metric is that it penalizes components in 
$\tilde{\mathbf{O}}$ that are not contained in the orientation representation $\tilde{\mathbf{R}}$ for symmetric object categories. To void penalizing the non-observable degrees of freedom, we select appropriate columns of $\tilde{\mathbf{O}}$ and construct all possible rotation matrices around them (e.g., matrices which represent rotations around z axis for \emph{bottles} and \emph{bowls}). We then compute ADD for each of the matrices and report the minimum ADD value.

Although in this way we take care of some of the ambiguities in symmetric objects, we also report the results with ADD-S metric~\cite{xiang2017posecnn}, where the average pairwise distance between the mesh surface points in the ground truth and estimated pose is computed using the closest point distance instead of the distance between corresponding points:
\begin{equation}
    \text{ADD-S} = \frac{1}{m} \sum_{\mathbf{x}_{m_1} \in \mathbf{M}} \operatorname*{min}_{\mathbf{x}_{m_2} \in \mathbf{M}}||(\mathbf{O}\mathbf{x}_{m_1} + \mathbf{p}) - (\tilde{\mathbf{O}}\mathbf{x}_{m_2} + \tilde{\mathbf{p}})||,
\end{equation}
where $\mathbf{x}_{m_1}, \mathbf{x}_{m_2} \in \mathbf{M}$ are mesh surface points.
This is because some objects might be symmetric only in a certain range of angle values, e.g., a handle of the mug might not be visible from a certain view. Selecting the $\tilde{\mathbf{O}}$ for which the ADD is minimum would not account for this. 

\subsection{Quantitative Evaluation}
\label{subsec:quanteval}
\subsubsection{Shape Retrieval}
For quantitative evaluation of shape retrieval, we compute the $\text{F1}$ score between $\mathbf{M_V}$ and $\tilde{\mathbf{M}}_\mathbf{V}$ (Fig.~\ref{fig:f1shape}). We also visualize the learned embedding space into which the test examples of the ``real" occluded dataset and the meshes from the training set are mapped (Fig.~\ref{fig:se}). The embedding shows that the similar image and mesh instances are mapped close to each other if similar. Since there are a lot of occlusions on the images in the test set, it is expected that some of the instances will not be close to any mesh, thus a gray area in the middle of the space. Quantitative results when training with and without the hand show slight improvement when hand configuration is added. 

\begin{figure*}
\centering
\scalebox{0.625}{
\includegraphics[width=\textwidth]{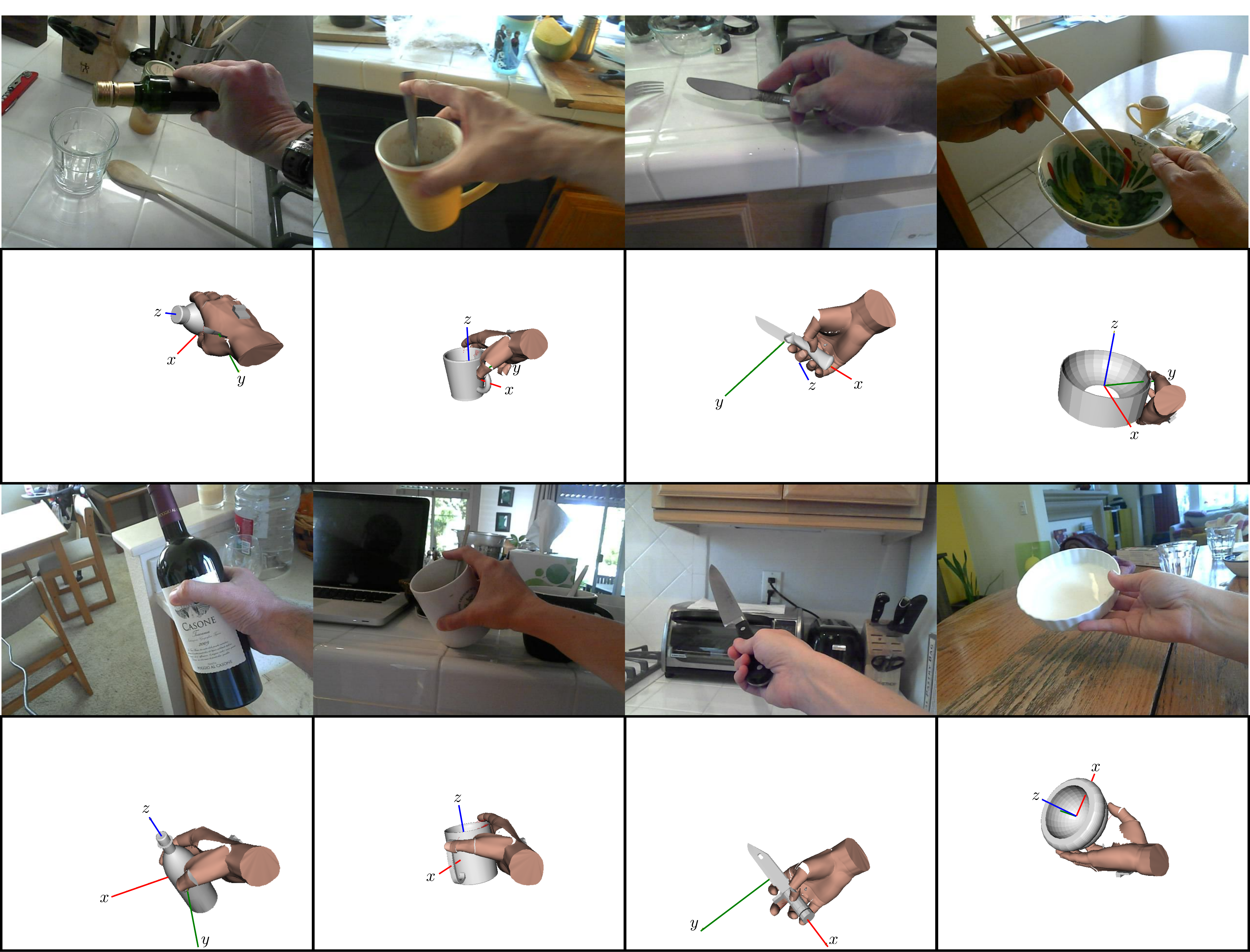}
}
\caption{Qualitative results on GUN71. We show two examples for \emph{bottle}, \emph{mug},  \emph{knife} and \emph{bowl} (from left to right). The first and the third row show images ``in the wild" and the second and the fourth row show the object pose and the hand pose and configurations after refinement.}
\label{fig:qual}
\end{figure*}

\begin{figure*}
\centering
\scalebox{0.625}{
\includegraphics[width=\textwidth]{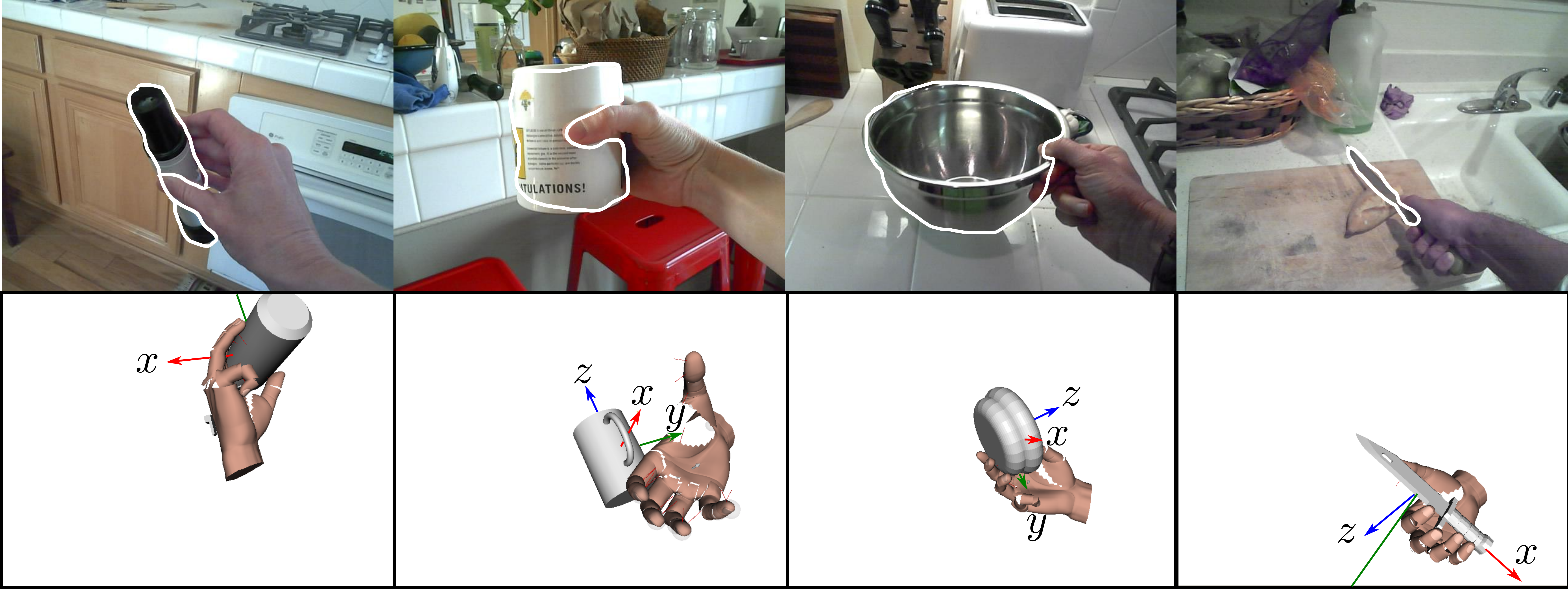}
}
\caption{Failure cases on GUN71. Top row shows images ``in the wild" (one for each category) and the bottom row shows the object pose and the hand pose and configuration after refinement. White outline on the images in the first row indicates results of \emph{Mask R-CNN} which affects the retrieved shape. Pose results show that when the estimates of the hand pose is wrong, the accuracy of the object pose is also incorrect.}
\label{fig:fail}
\end{figure*}

{\centering
\scalebox{0.85}{
\begin{minipage}{0.25\textwidth}
  \centering
  \includegraphics[width=6cm, height=6cm]{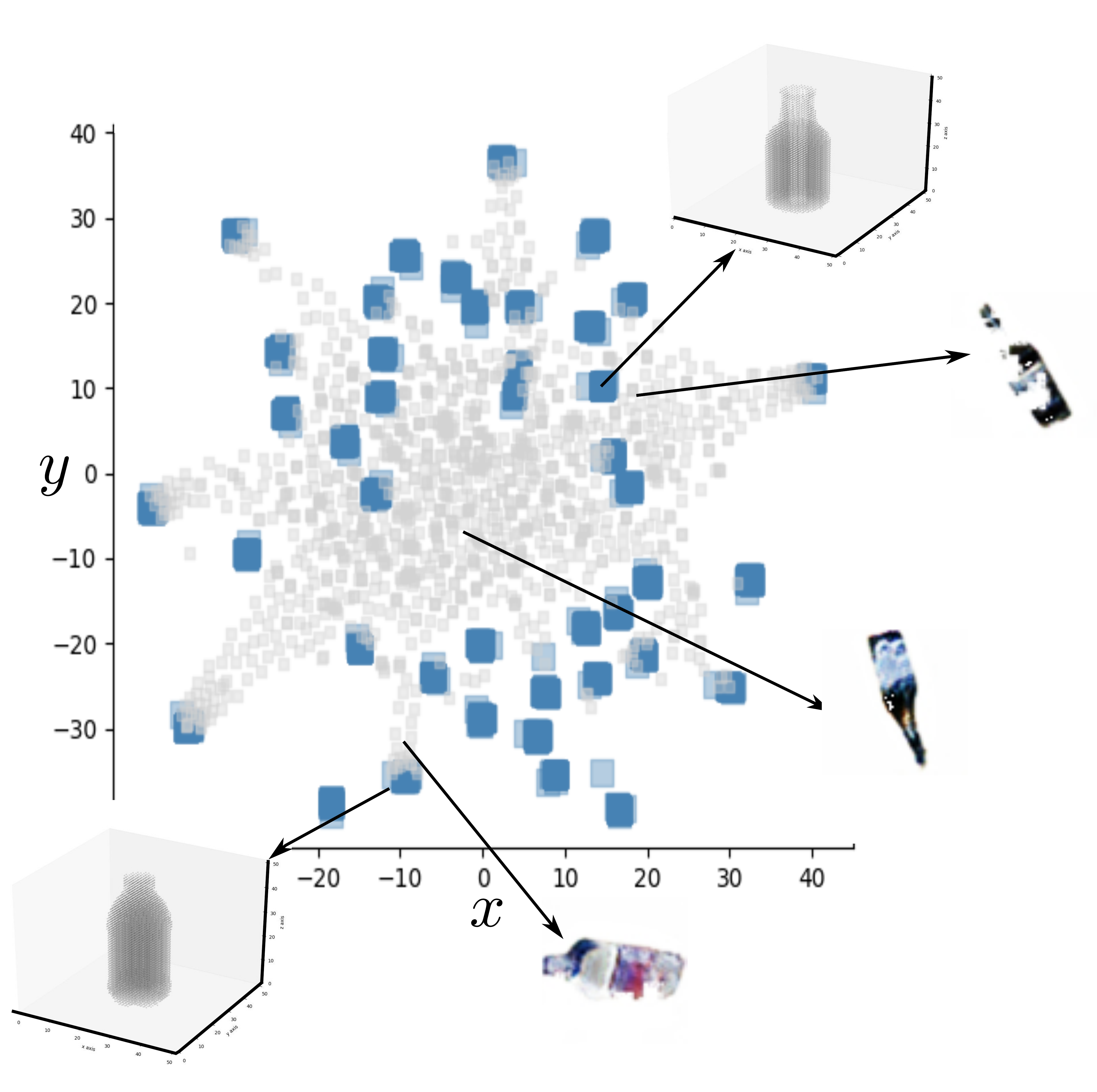}
  \captionsetup{justification=justified}
  \captionof{figure}{Shape embedding space. Blue squares indicate mapped mesh instances and gray squares image instances from the ``real" occluded test set.}
  \label{fig:se}
\end{minipage}
}
\hspace{1cm}
\scalebox{.8}{
\begin{minipage}{0.24\textwidth}
  \centering
    \begin{tabular}{ccc}
    \multicolumn{1}{l}{} & \multicolumn{2}{c}{$F_1$ score} \\ \hline
    $c$ & 
    \multicolumn{1}{c}{\begin{tabular}[c]{@{}c@{}}``real" \\occ.\end{tabular}} & \multicolumn{1}{c}{\begin{tabular}[c]{@{}c@{}}``real" \\occ. + \\$\mathbf{HP, HS}$\end{tabular}} \\ \hline
    bottle & 67.91 & 68.28 \\ 
    mug & 49.42 & 51.28 \\ 
    knife & 65.31 & 67.74  \\ 
    bowl & 41.35 & 43.77 \\ \hline
    \textbf{avg} & 55.99 & 57.76 
    \end{tabular}
    \captionsetup{justification=justified}
    \captionof{figure}{F1 scores for shape retrieval. The scores are calculated between $\tilde{\mathbf{M_V}}$ and $\mathbf{M_V}$ for training cases ii) and iii).}
    \label{fig:f1shape}
\end{minipage}
}
}

\subsubsection{Pose Estimation}
For the ablation study we consider four different cases of training data: i) ``real" images only, ii) ``real" occluded without the $\mathbf{HP, HS}$, iii) ``real" occluded with the noisy $\mathbf{HP, HS}$ and iv) ``real" occluded with the noisy $\mathbf{HP, HS}$ and refinement. We test it on the ``real" occluded set with the $\mathbf{HP, HS}$. The training and testing data are split $70: 30$ for each category.

Fig.~\ref{fig:curves} shows accuracy-threshold curves and Fig.~\ref{fig:abl} quantitative results for ADD and ADD-S metric for training cases ii) - iv). 
The results are better when training with the hand and even more improved after the refinement which confirms our hypothesis that the object pose estimation benefits from knowing the hand parameters. 
The improvements are expected if the grasps on the objects follow the distribution of real world grasps. We find empirically that when the grasps are generated randomly, there is no significant improvement in the shape and pose estimates.  
{\centering
\scalebox{0.65}{
\begin{minipage}{0.24\textwidth}
    \begin{tabular}{cccc}
    \multicolumn{1}{l}{} & \multicolumn{3}{c}{Position MAE [cm]} \\ \hline
    $c$ & 
    \multicolumn{1}{c}{\begin{tabular}[c]{@{}c@{}}``real" \\occ.\end{tabular}} & \multicolumn{1}{c}{\begin{tabular}[c]{@{}c@{}}``real" \\occ. + \\$\mathbf{HP, HS}$\end{tabular}} & \multicolumn{1}{c}{\begin{tabular}[c]{@{}c@{}}``real" \\occ. + \\$\mathbf{HP, HS}$ ref.\end{tabular}} \\ \hline
    bottle & 4.125 & 2.549 & 1.836\\ 
    mug & 3.957 & 3.188 & 2.936\\ 
    knife & 5.134 & 3.237 & 2.833\\ 
    bowl & 3.876 & 3.541 & 2.923\\ \hline
    \textbf{avg} & 4.277 & 3.076 & 2.686
    \end{tabular}
    \captionsetup{justification=justified}
    \captionof{figure}{Mean average error MAE between the $\mathbf{p}$ and $\tilde{\mathbf{p}}$  (in cm) for training cases ii)-iv).}
    \label{fig:pos}
\end{minipage}
}
\hspace{1.2cm}
\scalebox{0.65}{
\begin{minipage}{0.24\textwidth}
    \begin{tabular}{cccc}
    \multicolumn{1}{l}{} & \multicolumn{3}{c}{Orientation MAE [deg]} \\ \hline
    $c$ & 
    \multicolumn{1}{c}{\begin{tabular}[c]{@{}c@{}}``real" \\occ.\end{tabular}} & \multicolumn{1}{c}{\begin{tabular}[c]{@{}c@{}}``real" \\occ. + \\$\mathbf{HP, HS}$\end{tabular}} & \multicolumn{1}{c}{\begin{tabular}[c]{@{}c@{}}``real" \\occ. + \\$\mathbf{HP, HS}$ ref.\end{tabular}} \\ \hline
    bottle & 23.97 & 16.91 &  14.37 \\ 
    mug & 63.68 & 53.42 & 53.20 \\ 
    knife & 81.51 & 43.31 & 42.64 \\ 
    bowl & 30.50 & 28.35 & 27.60 \\ \hline
    \textbf{avg} & 49.91 & 35.49 & 34.45
    \end{tabular}
    \captionsetup{justification=justified}
    \captionof{figure}{Mean average error MAE between $\mathbf{O}$ and $\tilde{\mathbf{O}}$ (in \degree) for training cases ii)-iv).}
    \label{fig:ort}
\end{minipage}
}
}

Fig.~\ref{fig:pos} and Fig.~\ref{fig:ort} show mean average error (MAE) in position and orientation for training cases ii)-iv). Both the position and orientation estimates are better with the hand.
However, while the position is relatively easy to learn, i.e., the overall MAE is less than $5$cm, the orientation is much more challenging. The orientation results for \emph{bottle} and \emph{bowl} are better than for \emph{mug} and \emph{knife}. 
There are two reasons for that. First, for \emph{bottles} and \emph{bowls} the orientation representation is $\mathbf{R} \in \mathbb{R}^3$ whereas for \emph{mugs} and \emph{knives} $\mathbf{R} \in \mathbb{R}^{6 \times 1} \times \mathbb{R}^{3 \times 3}$. 
Second, \emph{mugs} and \emph{knives} require more distinctive features to be detected, e.g., the handle and opening in mugs and the lower side of the blade in knives. 
This is especially difficult if the aforementioned parts are not visible or occluded by the hand. \emph{Bowls} are also more challenging than the \emph{bottles}. 
This is because the concavity of the \emph{bowl}, which is the most informative of the orientation, is often visually ambiguous.

\subsection{Qualitative Evaluation on GUN71 dataset}
\label{subsec:qualeval}
Since there are no applicable real-world datasets annotated with object and hand poses, we provide only qualitative results on GUN-71 dataset~\cite{rogez2015understanding}.
We show two images per category from for which the \emph{HandNet} and \emph{Mask R-CNN} were successful, see Fig.~\ref{fig:qual}. The results show qualitatively reasonable object poses and shapes even in the case of large occlusions and transparency. The failure cases in Fig.~\ref{fig:fail} show that when the hand configuration is not correctly estimated, the accuracy of the estimated object pose degrades. This is because \emph{HOPS-Net} hinges on hand cues when there is uncertainty about the orientation, e.g., openings and bottoms of the bottles are sometimes difficult to discern and the correctly estimated hand configuration can help to resolve this. 
\section{Discussion and Conclusion}
\label{sec:conclusion}
We proposed an approach for 3D shape and 6D pose estimation of a novel hand-held object from an RGB image. We designed a CNN, \emph{HOPS-Net}, that takes as input the object segment together with a hand pose and configuration, and outputs a position, orientation and shape features used to retrieve the most similar mesh. It uses a hand to facilitate object shape and pose estimation and furthermore to refine the initial predictions and obtain plausible grasps in a grasping simulator.
The model was trained only on synthetic data which were made more realistic via an image-to-image translation network. 
The synthetic data also includes occlusions from the hand generated by grasping the objects with a model of the human hand in the simulator. 
The limitation of our method is that it hinges on existing approaches for pre-processing which are often inaccurate. In future, we plan on jointly learning to estimate hand and object. Furthermore, we plan on adding more object categories to scale the system.

{\small
\bibliographystyle{plainnat}
\bibliography{references}
}
\end{document}